\documentclass[conference,a4paper]{IEEEtran}
\IEEEoverridecommandlockouts
\usepackage{ifpdf}

\usepackage{cite}

\usepackage{amssymb}

\ifCLASSINFOpdf
\else
\fi
\usepackage[cmex10]{amsmath}
\usepackage{algorithmic}

\usepackage{array}

\usepackage{mdwmath}
\usepackage{mdwtab}

\usepackage{eqparbox}
\usepackage{fixltx2e}

\usepackage{graphicx}
\usepackage{multirow} 
\usepackage{stfloats}

\usepackage{url}

\begin{document}
%
\title{RoboPhoto-VLA: Robotic Photography System based on Vision-Language-Action Model}




%
\author{
    \IEEEauthorblockN{
        Song-Li Wu$^{1}$,
        Chubin Zhang$^{1}$,
        Wenbo Ding$^{1}$,
        Yansong Tang$^{1,*}$
    }
    \IEEEauthorblockA{
        $^{1}$Shenzhen International Graduate School,
        Tsinghua University
    }
    \IEEEauthorblockA{
        \{wsl24, zcb24\}@mails.tsinghua.edu.cn 
    }
    \IEEEauthorblockA{
    \{ding.wenbo, tang.yansong\}@sz.tsinghua.edu.cn
    }
    \thanks{This work was supported by the National Natural Science Foundation of China under Grant No.~U25A20537. $^{*}$Corresponding author.}
}


\maketitle

\begin{abstract}
Modern robotic photography requires understanding photographic intent, planning camera motions, and producing aesthetically pleasing images in complex environments. We present RoboPhoto-VLA, a unified Vision-Language-Action framework that jointly models perception, reasoning, and camera control for autonomous robotic photography. RoboPhoto-VLA incorporates PhotoCoT for structured action planning and Aesthetic Direct Preference Optimization (ADPO) to align camera behaviors with human aesthetic preferences. To facilitate training and evaluation, we construct OpenPhotography, a robotic photography dataset with language-guided demonstrations. Extensive real-world experiments demonstrate that RoboPhoto-VLA produces photographs better aligned with human aesthetic preferences while generalizing effectively to unseen subjects, viewpoints, poses, and complex photographic instructions.
\end{abstract}


%
\IEEEpeerreviewmaketitle

\section{Introduction}

Robotic photography is emerging as an important embodied AI application in creative production, event recording, and commercial media, where robots are expected to autonomously execute camera movements according to high-level user instructions~\cite{ma2024survey,lu2023thinkbot,zhang2026clap}. Unlike conventional robotic manipulation~\cite{lu2025vla,lu2025gwm,guo2026vla}, whose objective is successful task execution, robotic photography requires balancing two complementary objectives: accurately understanding the semantic intent of the user while producing photographs that satisfy human aesthetic preferences. Achieving this objective demands not only precise camera control, but also reasoning about abstract photographic concepts such as composition, viewpoint, framing, and subject interaction. Developing robots capable of jointly perceiving scenes, interpreting photographic intent, and generating aesthetically pleasing camera behaviors therefore remains a fundamental challenge~\cite{luo2026stare,lu2024manigaussian,wu2026activemem}.%

Existing robotic photography systems have explored reference-guided camera control, semantic event photography, and interactive computational photography~\cite{limoyo2024photobot,rivkin2023ansel,choi2024snapbot}. However, these methods are typically built upon modular pipelines that decouple perception, language understanding, and camera control~\cite{kawaharazuka2025vision,lin2025showui,adilkhanov2025survey}, limiting their ability to reason over abstract photographic intent and adapt to open-ended instructions, unseen subjects, and dynamic shooting environments. Recent Vision--Language--Action (VLA) models address this limitation by unifying perception, reasoning, and action within a single policy~\cite{nair2022r3muniversalvisualrepresentation,karamcheti2023languagedrivenrepresentationlearningrobotics,stone2023open,wu2026dpamixersr,wang2024autonomousbehaviorplanninghumanoid,wang2024hypermotion}. Nevertheless, existing VLAs are primarily optimized for task completion rather than photographic quality, lacking both photography-oriented reasoning and explicit optimization toward human aesthetic preferences. Moreover, the scarcity of embodied robotic photography datasets further limits their applicability to autonomous photography.

\begin{figure}[t]
  \centering
  \includegraphics[width=1\linewidth]{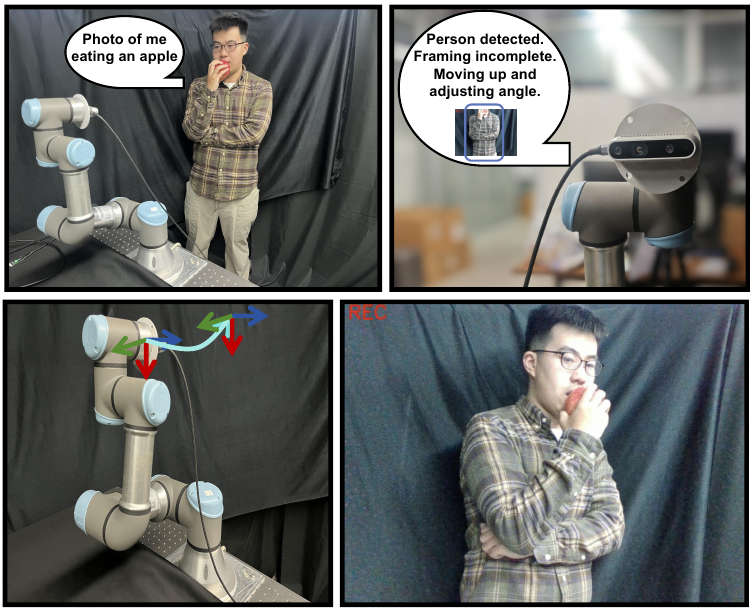}
  \caption{RoboPhoto-VLA first receives the user’s language request alongside a target scene (upper left: the user asks to capture a photo of themselves eating an apple). The system then detects the user, identifies incomplete framing of the pose, and initiates adjustments to its position and angle (upper right). Next, RoboPhoto-VLA precisely adjusts its robotic arm and camera orientation to align with the desired composition (lower left). The lower-right panel displays the final unretouched photograph captured by RoboPhoto-VLA.} 
  \label{fig1}
\end{figure}

To address these challenges, we formulate robotic photography as an aesthetic-aware Vision--Language--Action learning problem and propose RoboPhoto-VLA, a unified framework that jointly models perception, reasoning, and embodied camera control. Specifically, we introduce PhotoCoT, a photography-oriented reasoning strategy that grounds abstract photographic intent into structured camera operations, including viewpoint selection, framing adjustment, and camera motion planning. To further optimize photographic quality, we propose Aesthetical Direct Preference Optimization (ADPO), which explicitly aligns camera behaviors with human aesthetic preferences through preference learning rather than conventional task-success supervision. To facilitate training and evaluation, we further construct OpenPhotography, the first open-ended robotic photography dataset containing language-guided demonstrations together with aesthetic preference pairs.
Extensive real-world experiments demonstrate that RoboPhoto-VLA consistently produces photographs that better align with human aesthetic preferences while generalizing effectively across unseen subjects, viewpoints, poses, and diverse photographic scenarios.

Our contributions are threefold: we propose RoboPhoto-VLA for aesthetic-aware robotic photography, introduce PhotoCoT and ADPO for structured reasoning and aesthetic alignment, and construct OpenPhotography for real-world evaluation under diverse unseen settings.

\section{Related Works}
Recent Vision--Language Models (VLMs) have achieved remarkable progress in multimodal understanding by combining pretrained vision encoders with large language models, enabling strong performance on visual reasoning, captioning, and question answering tasks~\cite{hudson2019gqa,singh2019towards,li2023blip,laurenccon2024obelics,zhang2024vision,wu2026capedit}. Building upon these advances, Vision--Language--Action (VLA) models integrate perception, language, and robot control into a unified policy, allowing pretrained VLMs to directly predict robot actions for embodied manipulation and long-horizon planning~\cite{stone2023open,wang2026grpo,padalkar2023open,li2023vision,wu2026codeskill,karamcheti2024prismatic,kim2024openvla}. Representative models such as RT-2-X and OpenVLA demonstrate impressive generalization across robotic manipulation tasks but are primarily designed for task execution rather than photography-oriented reasoning and aesthetic decision-making.
Robotic photography has traditionally focused on camera control, viewpoint selection, framing, and target tracking through rule-based or modular pipelines~\cite{limoyo2024photobot,rivkin2023ansel,choi2024snapbot}. Although recent systems incorporate language understanding and scene perception, they generally lack unified reasoning over visual observations and abstract photographic intent, as well as explicit optimization toward human aesthetic preferences. In contrast, our work formulates robotic photography as an aesthetic-aware Vision--Language--Action problem. Built upon a unified VLA framework, RoboPhoto-VLA introduces photography-oriented reasoning and preference optimization to bridge abstract photographic intent with executable camera actions, enabling aesthetically aligned robotic photography.

\setcounter{figure}{1}
\begin{figure*}
\centering
\includegraphics[width=0.85\linewidth]{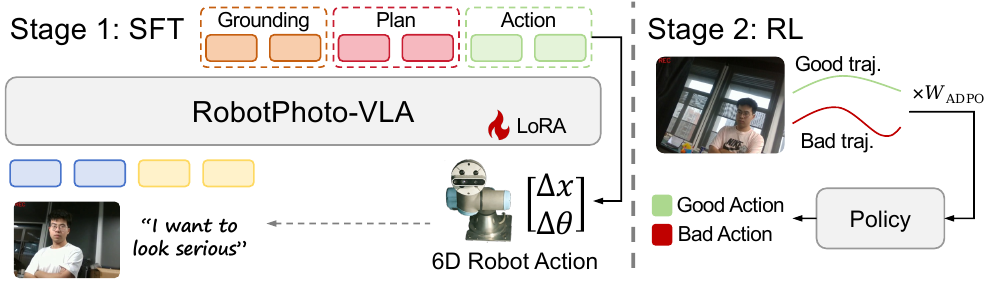}
\caption{RoboPhoto-VLA model architecture.
Given an image observation and a language instruction, the model predicts executable robot camera actions.
The architecture follows a two-stage training paradigm:
(1) Supervised Fine-Tuning (SFT) with PhotoCoT for structured photographic grounding and multi-step action planning;
(2) Reinforcement Learning (RL) with ADPO for refining action selection to better align with human aesthetic preferences.}
\label{framework}
\vspace{-10pt}
\end{figure*}

\section{Method}

We formulate robotic photography as an embodied vision--language--action (VLA) learning problem. Given an RGB observation $I$, a natural-language photography instruction $x$, and a camera action trajectory $A=\{a_1,\cdots,a_T\}$, the objective is to learn a policy
$\pi_\theta(A|I,x)$,
that not only executes feasible camera motions but also produces photographs aligned with human aesthetic preferences. To this end, we propose RoboPhoto-VLA, which extends a general-purpose VLA policy with photography-oriented reasoning and preference optimization. The overall framework consists of three components: an OpenVLA backbone for end-to-end policy learning, PhotoCoT for structured photographic reasoning, and Aesthetical Direct Preference Optimization (ADPO) for aesthetic alignment.

\subsection{OpenVLA-based Policy}

We adopt OpenVLA-7B as the backbone policy due to its strong embodied reasoning capability. Following OpenVLA, the policy consists of a dual visual encoder, a lightweight projector, and a large language model (LLM). Given an observation image, visual features are extracted by SigLIP and DINOv2,
$z_v=f_v(I)$,
while the language instruction is encoded as
$z_l=f_l(x)$.
The projected visual tokens and language tokens are concatenated and processed by the LLM to obtain a joint representation
$h=f_{\mathrm{LLM}}([z_v,z_l])$,
from which the policy autoregressively predicts the camera trajectory
$A\sim\pi_\theta(A|I,x)$.
Following OpenVLA, each continuous action dimension is independently discretized into 256 bins according to the 1\% and 99\% quantiles of the training distribution and represented as language tokens, allowing robot control to be optimized using the standard next-token prediction objective. Unlike OpenVLA, RoboPhoto-VLA explicitly introduces photography-oriented reasoning and aesthetic preference learning, enabling the policy to optimize photographic quality rather than only task completion.

\subsection{Photography-Oriented Chain-of-Thought}

General-purpose VLA models directly predict actions from observations, making it difficult to interpret abstract photographic instructions. We therefore introduce \textbf{PhotoCoT}, which decomposes policy learning into photographic grounding and camera planning.
Given an observation and instruction, PhotoCoT first generates a grounding representation
$g=f_g(I,x)=
\left\{
o,v,c,s
\right\}$,
where $o$, $v$, $c$, and $s$ denote the target subject, viewpoint, composition, and scene constraints, respectively. Instead of directly producing actions, the policy first predicts an intermediate reasoning sequence
$r=f_r(g)=
\{r_1,\cdots,r_K\}$,
which explicitly describes the multi-step camera adjustment process. The final camera trajectory is then generated as
$A=f_a(r)$.

\subsection{Aesthetic Direct Preference Optimization}

Although supervised fine-tuning enables reliable camera control, it does not explicitly optimize photographic aesthetics. We therefore formulate aesthetic alignment as a trajectory-level preference learning problem. For each observation--instruction pair $(I,x)$, multiple candidate camera trajectories are executed to capture corresponding images. These images are evaluated using a multimodal aesthetic assessment model, and the trajectories with higher and lower aesthetic scores are used to construct preference pairs $(y^{+},y^{-})$.
We initialize preference optimization from the PhotoCoT-supervised policy. The reference policy $\pi_{\mathrm{ref}}$ is a frozen copy of this policy, while $\pi_{\theta}$ denotes the policy optimized during preference learning. Following DPO, the preference margin is defined as
\begin{equation}
\Delta =
\beta
\left[
\log
\frac{\pi_{\theta}(y^{+}\mid I,x)}
     {\pi_{\mathrm{ref}}(y^{+}\mid I,x)}
-
\log
\frac{\pi_{\theta}(y^{-}\mid I,x)}
     {\pi_{\mathrm{ref}}(y^{-}\mid I,x)}
\right],
\end{equation}
where $\beta$ controls the strength of deviation from the reference policy. The standard DPO objective is
\begin{equation}
\mathcal{L}_{\mathrm{DPO}}
=
-
\mathbb{E}_{(I,x,y^{+},y^{-})}
\left[
\log \sigma(\Delta)
\right].
\end{equation}

Standard DPO treats all preference pairs equally, although their reliability may vary. We therefore introduce an adaptive confidence weight
\begin{equation}
w =
\sigma
\left(
\alpha\Delta_{\mathrm{aes}}
-
\gamma\Delta_{\mathrm{lpips}}
\right),
\end{equation}
where $\Delta_{\mathrm{aes}}$ denotes the aesthetic-score difference between the images produced by $y^{+}$ and $y^{-}$, and $\Delta_{\mathrm{lpips}}$ denotes their LPIPS perceptual distance. The coefficients $\alpha$ and $\gamma$ control the contributions of aesthetic confidence and perceptual variation, respectively. Consequently, preference pairs with a clear aesthetic margin receive larger weights, whereas pairs with large perceptual differences are down-weighted to reduce the influence of unrelated visual-content changes.

The resulting ADPO objective is
\begin{equation}
\mathcal{L}_{\mathrm{ADPO}}
=
-
\mathbb{E}_{(I,x,y^{+},y^{-})}
\left[
w\log \sigma(\Delta)
\right].
\end{equation}

The resulting optimization objective becomes
\begin{equation}
\mathcal{L}_{\mathrm{ADPO}}
=
-
\mathbb{E}
\left[
w
\log
\sigma
(
\Delta
)
\right].
\end{equation}

Finally, RoboPhoto-VLA is trained using a two-stage paradigm. We first perform supervised fine-tuning with PhotoCoT reasoning traces to learn photography-oriented grounding and camera planning. The resulting policy is subsequently optimized using ADPO to align camera trajectories with human aesthetic preferences. The overall objective is
\begin{equation}
\mathcal{L}
=
\mathcal{L}_{\mathrm{SFT}}
+
\lambda
\mathcal{L}_{\mathrm{ADPO}},
\end{equation}
where $\lambda$ balances supervised policy learning and preference optimization. During inference, RoboPhoto-VLA directly predicts an executable camera trajectory from an RGB observation and a natural-language instruction through a single forward pass.

\section{Experiments}


\subsection{Experimental Setup}
All experiments are conducted on a UR-series robotic arm equipped with an Intel RealSense D435i RGB-D camera. The balancing coefficient $\lambda$ is set to 0.1 based on validation performance, balancing camera-control capability and aesthetic alignment.
OpenPhotography contains 600 demonstrations covering 42 instructions, diverse subjects, viewpoints, camera heights, and human poses. Each demonstration includes an RGB observation, an instruction, a manually verified Qwen3-VL-generated PhotoCoT trace~\cite{bai2025qwen3}, a camera trajectory, and the resulting photograph. The demonstrations are split into training, validation, and test sets at a ratio of 6:2:2.
For OOD evaluation, samples are further divided into in-distribution and OOD subsets at a ratio of 7:3, with all OOD samples excluded from training and preference optimization. Cross-subject evaluation separately considers unseen single-person and multi-person scenes; cross-view evaluation uses $45^\circ$, $90^\circ$, and $135^\circ$ viewpoints; and cross-posture evaluation considers unseen human poses. For preference learning, multiple SFT-policy trajectories are executed and their photographs are scored by a multimodal aesthetic assessment model~\cite{du2026venus}. After removing failed or ambiguous comparisons, 200 trajectory-level preference pairs are retained, with the higher- and lower-scoring trajectories labeled $y^{+}$ and $y^{-}$, respectively.

We compare RoboPhoto-VLA with three representative general-purpose VLA models (RT-2-X, $\pi_0$, and OpenVLA) and two representative robotic photography systems (ANSEL and Snapbot). The VLA baselines are fine-tuned on OpenPhotography using their official implementations, while ANSEL and Snapbot are evaluated using their original photography pipelines.
Following prior work, we report three complementary metrics. Quality Score (QS)~\cite{wu2023q} measures overall image quality, including sharpness, exposure, and framing. Aesthetic Score (AS)~\cite{huang2024aesexpert} evaluates photographic composition and visual attractiveness using a pretrained aesthetic assessment model. Instruction--Image Alignment (IIA)~\cite{radford2021learning} computes CLIP similarity between the generated image and the corresponding language instruction. All reported numbers are averaged over the testing set, and higher values indicate better performance.

\begin{table*}[t]
\centering
\caption{Performance comparison of RoboPhoto-VLA with other models for photography tasks. "HoP" denotes Human-only Photography and "HOP" denotes Human–Object Photography}
\resizebox{1.\textwidth}{!}{%
\begin{tabular}{c| c|cc c|cc c|cc c|cc c|cc c|ccc}
\hline 
\multirow{2}{*}{\textbf{Experiment}} & \multirow{2}{*}{\textbf{Subtask}} &  \multicolumn{3}{c|}{\textbf{RT-2-X} }& \multicolumn{3}{c|}{\textbf{$\pi_0$}} & \multicolumn{3}{c|}{\textbf{OpenVLA}} & \multicolumn{3}{c|}{\textbf{ANSEL}} & \multicolumn{3}{c|}{\textbf{Snapbot}} & \multicolumn{3}{c}{\textbf{RoboPhoto-VLA}} \\
\cline{3-20}
& & \textbf{QS} &\textbf{AS} & \textbf{IIA} & \textbf{QS} &\textbf{AS} & \textbf{IIA} & \textbf{QS} &\textbf{AS} & \textbf{IIA}& \textbf{QS} &\textbf{AS} & \textbf{IIA}&\textbf{QS} &\textbf{AS} & \textbf{IIA}& \textbf{QS} &\textbf{AS} & \textbf{IIA}\\
\hline
\multirow{2}{*}{Cross-subject}  
 & Single-Person &0.09 & 0.05 & 0.02   & 0.37 & 0.48 & 0.35 & 0.29 & 0.41 & 0.32  &0.63 & \underline{0.78} & 0.65  & \underline{0.72} & 0.61 & \underline{0.69}  & \textbf{0.86} & \textbf{0.91} & \textbf{0.87} \\
 \cline{2-20}
 & Multi-Person& 0.03 & 0.07 & 0.01  &  0.21 & 0.31 & 0.32 & 0.27 & 0.36 & 0.38  &  \underline{0.75} & 0.66 & \underline{0.71}  &  0.64 & \underline{0.77} & 0.68  & \textbf{0.89} & \textbf{0.90} & \textbf{0.91} \\ 
\hline
\multirow{3}{*}{Cross-view} 
 & 45° & 0.03 & 0.04 & 0.06  & 0.22 & 0.46 & 0.30 & 0.21 & 0.23 & 0.37  &  0.62 & 0.73 & \underline{0.70}  & \underline{0.67} & \underline{0.76} & 0.65  &\textbf{0.73} & \textbf{0.85} & \textbf{0.89}\\ 
 \cline{2-20}
 & 90° &0.07 & 0.08 & 0.09  &  0.35 & 0.49 & 0.36 & 0.20 & 0.45 & 0.39  &\underline{0.74} & 0.63 & \underline{0.78}  &  0.66 & \underline{0.71}& 0.69 & \textbf{0.88} & \textbf{0.81} & \textbf{0.86} \\  \cline{2-20}
 & 135° &0.00 & 0.03 & 0.03 & 0.24 & 0.40 & 0.31 &0.22 & 0.31 & 0.30  & 0.64 & \underline{0.75} & 0.62  & \underline{0.73} & 0.70 & \underline{0.67} & \textbf{0.85} & \textbf{0.83} & \textbf{0.85}\\ 
\hline
\multirow{2}{*}{Cross-Posture} 
 & HoP &0.05 & 0.06 & 0.08  &0.38 & 0.42 & 0.33 &0.25 & 0.48 & 0.35  & \underline{0.76} & \underline{0.68} & 0.65  & 0.72 & 0.63 & \underline{0.77} & \textbf{0.83} & \textbf{0.86} & \textbf{0.84}\\  \cline{2-20}
 & HOP & 0.04 & 0.02 & 0.00 & 0.26 & 0.47 & 0.34 & 0.29 & 0.44 & 0.32  &  \underline{0.66} & 0.71 & \underline{0.69}  & 0.64 & \underline{0.74} & 0.62 &  \textbf{0.74} & \textbf{0.80} & \textbf{0.73} \\ 
\hline 
\end{tabular}
}
\label{tab.Performance_results}
\end{table*}

\subsection{Comparison with State-of-the-Art Methods}

Table~\ref{tab.Performance_results} reports the comparison with representative VLA models and robotic photography systems. RoboPhoto-VLA consistently achieves the best performance across all cross-subject, cross-view, and cross-posture settings. Compared with general-purpose VLAs (RT-2-X, $\pi_0$, and OpenVLA), our method significantly improves both aesthetic quality (AS) and instruction alignment (IIA), demonstrating the importance of photography-oriented reasoning and preference alignment. Compared with traditional photobots (ANSEL and Snapbot), RoboPhoto-VLA further produces higher-quality images while preserving accurate instruction following. The largest gains are observed in multi-person scenes and challenging viewpoints, where structured PhotoCoT reasoning and ADPO jointly enable robust subject grounding, multi-step camera planning, and aesthetically consistent image capture.

\subsection{Cross-Domain Generalization}

We further evaluate out-of-distribution generalization on unseen subjects, viewpoints, and human poses. As shown in Table~\ref{tab.Performance_results}, existing VLA models exhibit noticeable performance degradation under all distribution shifts, particularly for unseen identities and viewpoints. In contrast, RoboPhoto-VLA consistently maintains high QS, AS, and IIA scores across all settings. The results indicate that PhotoCoT learns transferable photographic reasoning beyond subject appearance, while ADPO improves the robustness of camera planning and aesthetic composition under diverse real-world scenarios.

\begin{figure}[t]
  \centering
  \includegraphics[width=0.85\linewidth]{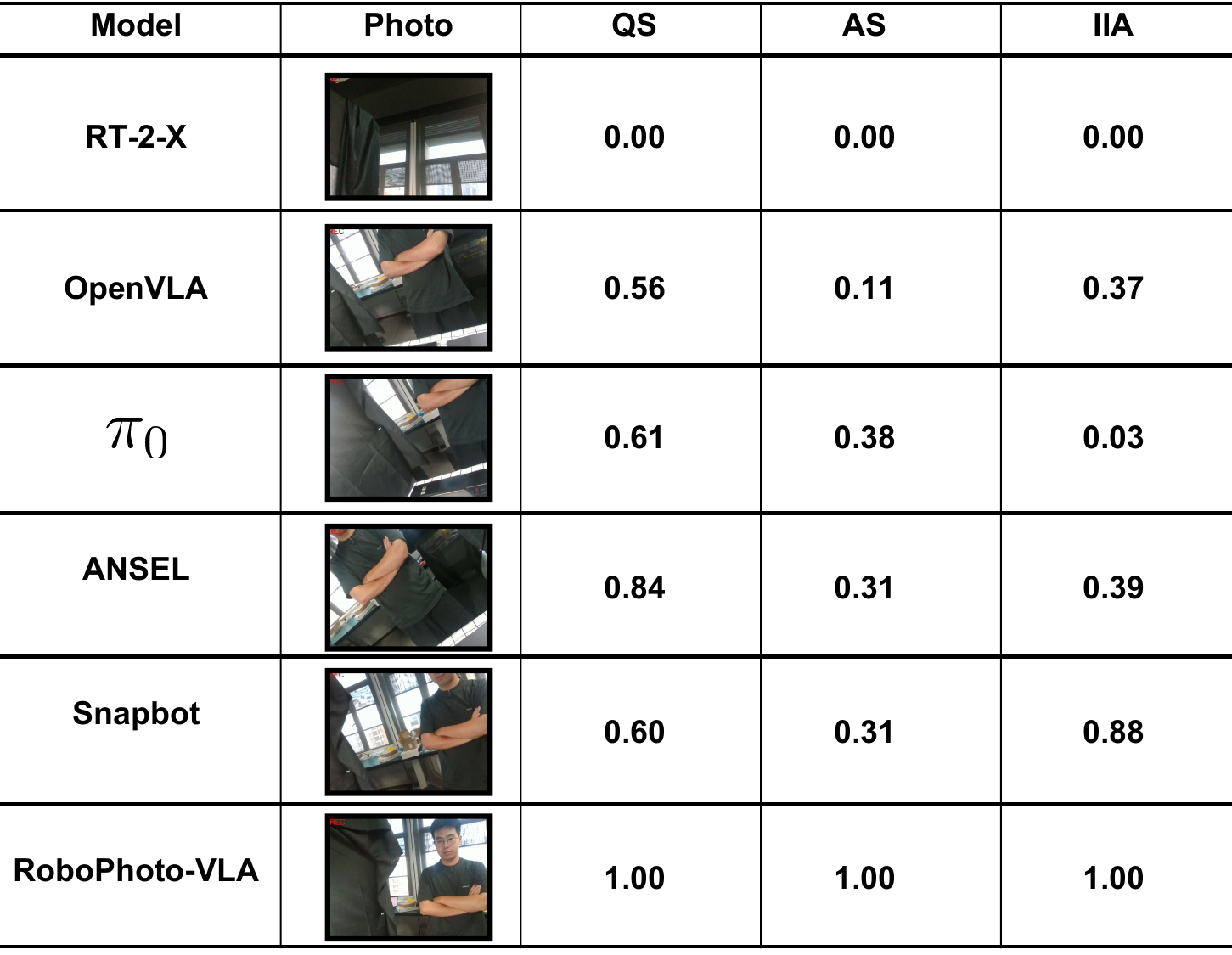}
  \caption{Instruction: Photograph me looking serious.} 
  \label{case}
\end{figure}

\begin{table*}[t]
\centering
\caption{Ablation Study}
\resizebox{0.95\textwidth}{!}{%
\begin{tabular}{c| c|ccc|ccc|cc c|cc c|ccc}
\hline 
\multirow{2}{*}{\textbf{Experiment}} & \multirow{2}{*}{\textbf{Subtask}} & \multicolumn{3}{c|}{\textbf{OpenVLA} } & \multicolumn{3}{c|}{\textbf{w/ SFT} }& \multicolumn{3}{c|}{\textbf{w/ ADPO}} &  \multicolumn{3}{c|}{\textbf{w/ DPO}} & \multicolumn{3}{c}{\textbf{RoboPhoto-VLA}} \\
\cline{3-17}
& &  \textbf{QS} &\textbf{AS} & \textbf{IIA} &  \textbf{QS} &\textbf{AS} & \textbf{IIA} &  \textbf{QS} &\textbf{AS} & \textbf{IIA}&\textbf{QS} &\textbf{AS} & \textbf{IIA}& \textbf{QS} &\textbf{AS} & \textbf{IIA}\\
\hline
\multirow{2}{*}{Cross-subject}  
 & Single-Person & 0.29 &0.41 & 0.32  & 0.53 & 0.57 & 0.47 & 0.73 & 0.81 & 0.74 &  0.67 & 0.61 & 0.65 &\textbf{0.76} & \textbf{0.84} & \textbf{0.79}\\
 \cline{2-17}
 & Multi-Person& 0.27 & 0.36 & 0.38 &0.38 & 0.42 & 0.56 & 0.74 & 0.72 & 0.75 &  0.68 & 0.52 & 0.63 &\textbf{0.81} & \textbf{0.83} & \textbf{0.84} \\
\hline
\multirow{3}{*}{Cross-view} 
 & 45°  &0.21 &0.23 &0.37 &0.26 & 0.52 & 0.41 & 0.63 & 0.72 & 0.81 &  0.56 & 0.53 & 0.69 &\textbf{0.73} & \textbf{0.85} & \textbf{0.89}\\  \cline{2-17}
 & 90° &0.20 &0.45 &0.39 & 0.46 & 0.62 & 0.44 & 0.78 & 0.77 & 0.73 & 0.71 & 0.63 & 0.66 & \textbf{0.88} & \textbf{0.81} & \textbf{0.86} \\  \cline{2-17}
 & 135° & 0.22 &0.31 &0.30  &0.31 & 0.58 & 0.42 & 0.71 & 0.75 & 0.78 &  0.63 & 0.55 & 0.63 & \textbf{0.85} & \textbf{0.83} & \textbf{0.85}\\ 
\hline 
\end{tabular}
}
\label{tab.ablation}
\end{table*}

\subsection{Qualitative Results}

\begin{figure}[t]
  \centering
  \includegraphics[width=1\linewidth]{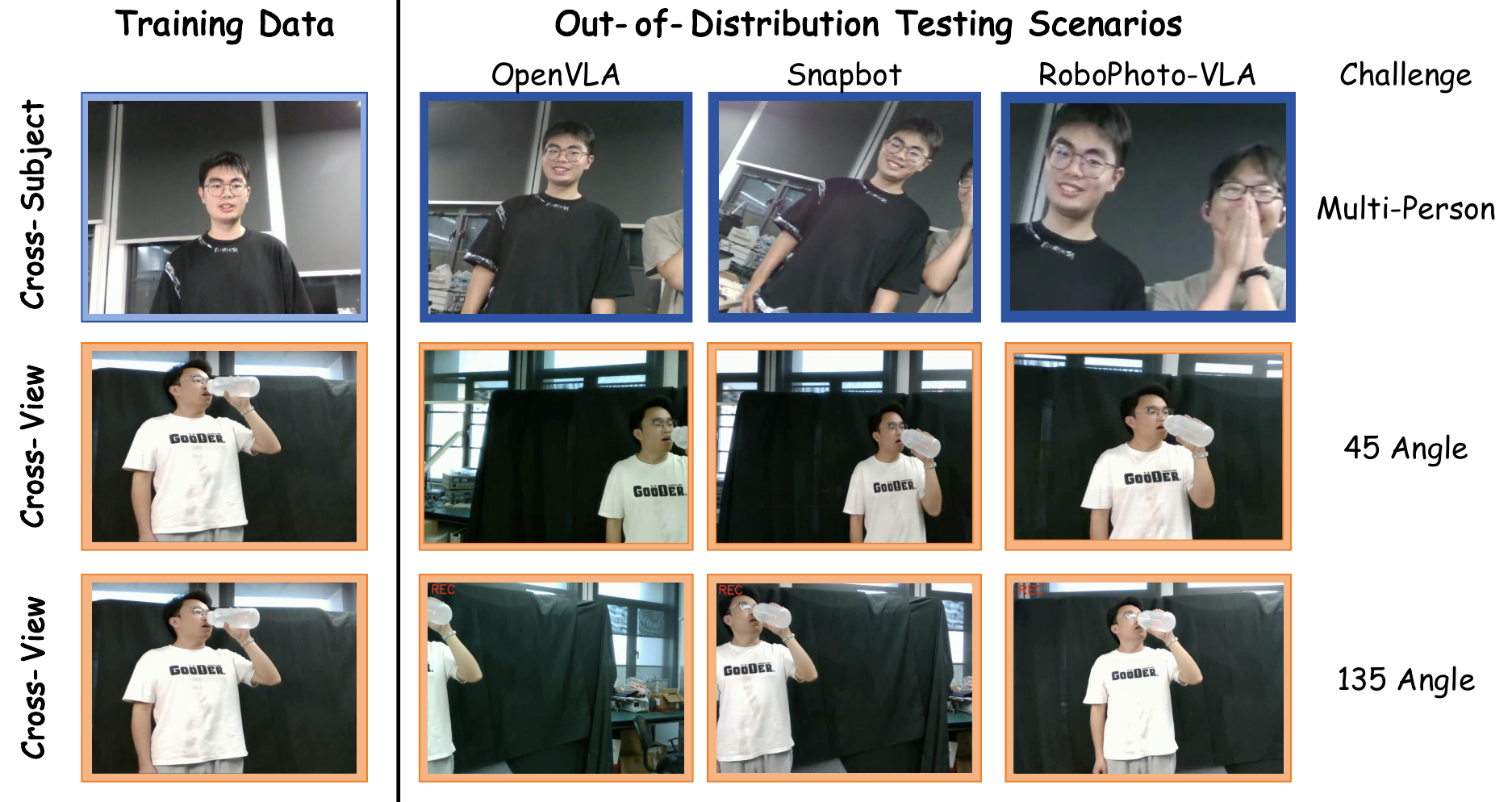}
  \caption{
We evaluate the model under cross-subject and cross-view generalization settings.
For cross-subject evaluation, we employ the instructions 
``Capture me smiling naturally'' and 
``Photograph me laughing with a friend'' 
to assess performance in single-person and multi-person scenarios, respectively. 
For cross-view evaluation, we use the instruction 
``Take a photo of me drinking water'' 
to examine robustness to viewpoint variations.
}
  \label{ODV}
\end{figure}

Representative results are shown in Fig.~\ref{case}. Compared with OpenVLA and Snapbot, RoboPhoto-VLA better understands photographic intent, producing more accurate framing, viewpoint selection, and facial expression. As illustrated in Fig.~\ref{ODV}, RoboPhoto-VLA also generalizes well to unseen subjects and viewpoints, generating consistently coherent and aesthetically pleasing photographs.

\subsection{Ablation Study}

Table~\ref{tab.ablation} summarizes the ablation results. Removing PhotoCoT leads to the largest drop in IIA, demonstrating the importance of structured photographic reasoning. Replacing ADPO with supervised fine-tuning or standard DPO mainly reduces AS, confirming that ADPO effectively aligns camera actions with human aesthetic preferences. Combining PhotoCoT and ADPO consistently achieves the best overall performance.

\section{Conclusion}

We present RoboPhoto-VLA, a unified vision--language--action framework for autonomous robotic photography that bridges high-level photographic intent and low-level camera control. By combining PhotoCoT with aesthetic preference optimization via ADPO, our framework enables robots to reason about composition, plan coherent camera trajectories, and capture images aligned with human visual preferences within a unified policy.
Extensive real-world experiments across diverse subjects, viewpoints, and poses demonstrate consistent improvements over general-purpose VLA models and existing robotic photography systems. We also introduce OpenPhotography, an open benchmark for advancing research in embodied robotic photography. We hope this work establishes robotic photography as a valuable testbed for embodied intelligence and advances research on aesthetic reasoning, human--robot creative collaboration, and preference-aware action learning.

\bibliographystyle{IEEEtran}
\bibliography{references}

\end{document}